\documentclass[runningheads]{llncs}

\usepackage{amsmath}
\usepackage{amssymb}
\usepackage[latin1]{inputenc}           
\usepackage[T1]{fontenc}                
\usepackage{epsfig}
\usepackage{color}
\usepackage{algorithm}
\usepackage{algorithmic}
\usepackage{subfigure}

\newcommand{\ZZ}{\mbox{\rm \lower0.3pt\hbox{$\angle\!\!\!$}Z}}

\newcommand{\dartset}{\mbox{$\mathcal{D}$}}

\newcommand{\Equ}[1]{Eq.~(\ref{#1})}

\newcommand{\Assign}{\ensuremath{\leftarrow}}

\newcommand{\DC}{\ensuremath{C}}
\newcommand{\PT}[1]{\ensuremath{{\DC}_{#1}}}
\newcommand{\PTS}[2]{\ensuremath{{\DC}_{#1,#2}}}
\newcommand{\SPRED}[2]{\ensuremath{S(#1,#2)}}

\newcommand{\LF}{\ensuremath{\lambda}}
\newcommand{\LTO}[1]{\ensuremath{\hat{\theta}(#1)}}

\newcommand{\LE}[1]{\ensuremath{\hat{l}(#1)}}

\newcommand{\DD}[1]{\ensuremath{\theta(#1)}}

\newcommand{\NORMAL}[1]{\ensuremath{\hat{\mathbf{n}}(#1)}}
\newcommand{\LENGTH}[2]{\ensuremath{\hat{L}(#1;#2)}}
\newcommand{\PERIM}[2]{\ensuremath{\widehat{\mathrm{Per}}(#1;#2)}}

\newcommand{\E}[0]{\ensuremath{E}}
\newcommand{\Eimg}[0]{\ensuremath{E_\text{img}}}
\newcommand{\Ereg}[0]{\ensuremath{E_\text{reg}}}

\newcommand{\Etal}[0]{\textit{et. al.}}

\newcommand{\dual}[1]{\mbox{\ensuremath{\overline{#1}}}}
\graphicspath{{Fig/},{FIGURES/},{FigMartin/}}

\mainmatter
\title{Combinatorial pyramids and discrete geometry for energy-minimizing segmentation}
\titlerunning{Lecture Notes in Computer Science}

\author{Martin Braure de Calignon$\dag$ \and Luc Brun$\ddag$ \and Jacques-Olivier Lachaud$\dag$}

 \institute{%
  \begin{minipage}{5cm}
    \centering
    $\dagger$LaBRI CNRS UMR 5800\\
    Université Bordeaux 1\\
    351, cours de la Libération\\
    33405 Talence cedex \\
    \email{\{braure,lachaud\}@labri.fr}
  \end{minipage}
  \begin{minipage}{5cm}
    \centering
    $\ddagger$GreyC CNRS UMR 6072\\
    {É}quipe Image - ENSICAEN\\
    6, Boulevard du Mar{é}chal Juin\\
    14050 CAEN Cedex - France\\
    \email{\{luc.brun\}@greyc.ensicaen.fr}
  \end{minipage}
}
\date{}

\begin{document}
\maketitle 

\begin{abstract}
  This paper defines the basis of a new hierarchical framework for
  segmentation algorithms based on energy minimization schemes.  This
  new framework is based on two formal tools.  First, a combinatorial
  pyramid encode efficiently a hierarchy of partitions.  Secondly,
  discrete geometric estimators measure precisely some important geometric
  parameters of the regions. These measures combined with
  photometrical and topological features of the partition allows to
  design energy terms based on discrete measures. Our segmentation
  framework exploits these energies to build a pyramid of image
  partitions with a minimization scheme. Some experiments illustrating
  our framework are shown and discussed.
\end{abstract}


\section{Introduction}
\label{sec:intro}

The convergence of energy minimization and hierarchical segmentation
algorithms provides a rich framework for image segmentation.  This
framework is based on an objective criterion, called {\em energy},
whose minimization defines a salient partition according to a given
problem .  The energy of a partition is generally decomposed by
summation over each region as a weighted sum of two terms
\( E(R)=E_{img}(R)+\nu E_{reg}(R) \) where $E_{img}$ may be understood
as a fit to the data within the region while $E_{reg}$ corresponds to
a regularization term. The parameter $\nu$ defines the respective weights
of the two terms. The Mumford-Shah energy is a classical instance of
this approach~\cite{chan-01}. Such equation may also be
interpreted within the Minimum Description Length (MDL)
framework~\cite{Leclerc89}, where the two energies $E_{img}$ and
$E_{reg}$ represent respectively the encoding costs of the photometry
and the geometry of a region.

Several methods have been proposed in order obtain a partition
minimizing an energy. These methods include the level set
approach~\cite{chan-01}, graph cuts~\cite{boykov-04} and the methods
based on a region merging
scheme~\cite{koepfler-94,Morel95,reddings-99,guigues-06}.  The
definition of a meaningful segmentation using an energy minimization
framework and a merge scheme supposes first to define a merge
strategy. If the parameter $\nu$ is fixed, a near optimal strategy
consists to merge at each step the two regions, the merging of which
induces the greatest decrease of the energy until any merge would
increase the energy. The obtained partition is said to be {\em 2
normal} at the scale $\nu$~\cite{koepfler-94,Morel95}.  An alternative
strategy~\cite{reddings-99} consists to merge at each step the two
regions whose union would belong to the 2 normal partition of lowest
scale.  This reduction framework avoids the need to select a vector of
$\nu$ parameters encoding {\em a priori} the different scales of
interest.  However, previous
works~\cite{koepfler-94,Morel95,reddings-99} where based on a sequence
of merge operations combined with a stopping criterion (number of
regions, maximal value of $\nu$\ldots).
Guigues~\Etal~\cite{guigues-06} encode explicitly the hierarchy of
partitions using a reduction scheme similar to~\cite{reddings-99} but
uses the hierarchy in order to build for any value of $\nu$, the
optimal partition which may be defined from the hierarchy. Moreover,
instead of starting from the grid of pixels like \cite{reddings-99},
their initial partition is an over partition of the image, which
presents two fundamental advantages. First, the initial over segmented
partition allows to compute reliable statistics on regions.  Secondly,
it restricts the set of possible partitions and thus reduces the risk
to be trapped into a local minima.

The second problem that should be addressed by a segmentation
algorithm is the correct design of the energy terms. For instance, the
classical Mumford-Shah energy simply combines the squared error of
each region together with the total length of the partition
boundaries. However, as shown by several authors~\cite{guigues-06},
more complex models (both geometrical and photometrical) may handle
finer definitions of salient partitions. Their design requires to fit
geometrical models onto regions. An efficient access to the set of
boundaries of each region and to their geometry is thus compulsory.
However, classical hierarchical segmentation frameworks are not
adequate for this task.  Adaptive pyramids based on
graph~\cite{jolion-01} do not present a 1-1 correspondence between
region adjacencies and geometrical boundaries: reconstructing the
geometry of a region is then tricky. Dual graphs~\cite{kropatsch03}
behave better for this task but the explicit encoding of all reduced
graphs restricts the number of merge steps.

This paper provides a new framework that addresses the design of new
energy terms based on geometrical and photometrical features. The
stack of successively reduced partitions is encoded using a
combinatorial pyramid~\cite{brun-06-1}. A very fine granularity for
the hierarchy is then achieved since regions are merged two by two and
a new level of the pyramid is created for each merging operation.
Geometrical features are computed on each partition of the hierarchy
using discrete geometric estimators of normal and length.  This
framework offers then a compact and efficient encoding of the
hierarchy together with an efficient access to the geometrical and
topological properties of the partition. It came thus as a natural
complement to methods searching for optimal partitions. The paper is
structured as follows. We present in Section~\ref{sec:combi_pyr} the
combinatorial pyramid model. The application of this model to compute
geometrical features on regions using discrete geometric estimators is
presented in Section~\ref{sec:geometry}.  We then present in
Section~\ref{sec:energy} one energy based on discrete estimators
together with some experiments.


\section{Combinatorial Pyramids}
\label{sec:combi_pyr}

This paper is based on combinatorial maps~\cite{lienhardt-91}. A
combinatorial map may be seen as a planar graph encoding explicitly
the orientation of edges around a given vertex. To do so, each edge of
a planar graph is split into two half-edges called {\em darts} (e.g.
darts $16$ and $-24$ in Fig.~\ref{fig:grid}c). Since each edge
connects two vertices, each dart belongs to only one vertex. A
combinatorial map is formally defined by a triplet $G=(\dartset,\sigma,\alpha)$
where $\dartset$ represents the set of darts and $\sigma$ is a permutation
on $\dartset$ whose cycles correspond to the sequence of darts
encountered when turning counter-clockwise around each vertex. Finally
$\alpha$ is an involution on $\dartset$ which maps each of the two darts of
one edge to the other one (e.g.  $\alpha$ maps $16$ to $-24$ and $-24$ to
$16$ in Fig~\ref{fig:grid}c).  The cycles of $\alpha$ and $\sigma$ containing a
dart $d$ will be respectively denoted by $\alpha^*(d)$ and $\sigma^*(d)$.

Given a combinatorial map $G=(\dartset,\sigma,\alpha)$, its dual map is defined
by $\dual{G}=(\dartset,\varphi,\alpha)$ with $\varphi=\sigma\circ\alpha$. The cycles of permutation
$\varphi$ encode the faces of the combinatorial map and may be interpreted
as the sequence of darts encountered when turning clockwise around a
face.  The cycle of $\varphi$ containing a dart $d$ will be denoted by
$\varphi^*(d)$.

\subsection{Combinatorial map encoding of a planar sampling grid}
\label{subsec:sampling}

\begin{figure}[t]
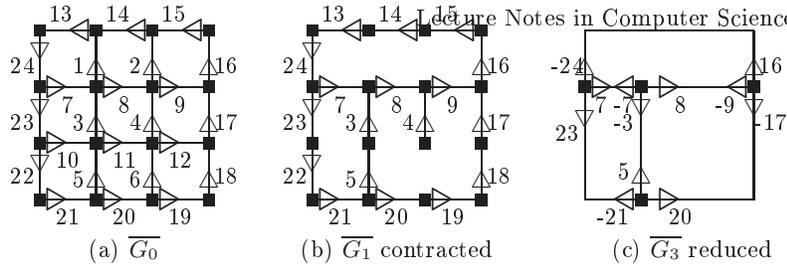

  \input{nodes.pic} \input{brins.tex}

  \unitlength 0.3mm 
\mbox{ }\hfill
\subfigure[\dual{G_0}]{
\begin{picture}(100.00,65.00)
  \input{grille_dual.pin}
\end{picture}
}\hfill
\subfigure[\dual{G_1} contracted]{
\begin{picture}(100.00,65.00)
  \input{grille_dual_cont.pin}
\end{picture}
}\hfill
\subfigure[$\dual{G_3}$ reduced]{
        \begin{picture}(120,65)
          \input{removed_grille_dual.pin}
        \end{picture}
}
\hfill\mbox{ }

\caption{A dual of a combinatorial map (a) encoding a $3\times 3$ grid with
  the contracted combinatorial map (b) obtained by the contraction of
  the contraction kernel (CK) $K_1=\alpha^*(1,2,10,11,12,6)$. The reduced
  combinatorial map (c) is obtained by the removal of the empty self
  loops defined by the RKESL $K_2=\alpha^*(4)$ and the removal kernel of
  empty double edges (RKEDE)
  $K_3=\alpha^*(13,14,15,19,18,22)\cup\{24,-16,17,-20,21,-23,3,-5\}$.}
\label{fig:grid}
\end{figure}

Combinatorial maps can also code the low level geometry of image
pixels. Indeed, Fig.~\ref{fig:grid}a describes a dual combinatorial
map $\dual{G_0}=( \dartset_0,\varphi_0,\alpha_0)$ encoding a $3\times 3$ 4-connected
planar sampling grid. The $\varphi$, $\alpha$ and $\sigma$ cycles of each dart may be
respectively understood as elements of dimensions $0$, $1$ and $2$ and
formally associated to a 2D cellular complex~\cite{brun-06-1}.  More
precisely, each $\alpha_0$ cycle may be associated to a {\em linel}
(sometimes also called {\em crack} or {\em surfel}) between two
pixels. Each of the two darts of an $\alpha_0$ cycle corresponds to an
orientation along the linel. For example, the cycle $\alpha_0^*(1)=(1,-1)$
is associated to the linel encoding the right border of the top left
pixel of the $3\times 3$ grid (Fig.~\ref{fig:grid}a). Darts $1$ and $-1$
define respectively a bottom to top and top to bottom orientation
along the linel.

\subsection{Construction of Combinatorial Pyramids}

A combinatorial pyramid is defined by an initial combinatorial map
successively reduced by a sequence of contraction or removal
operations. Contraction operations are encoded by contraction kernels
(CK).  These kernels, defined as a forest of the current combinatorial
map, may however create redundant edges such as empty-self loops and
double edges. Empty self loops (edge $\alpha_1^*(4)$ in
Fig.~\ref{fig:grid}b) may be interpreted as region inner boundaries
and are removed by a removal kernel of empty self loops (RKESL) after
the contraction step. The remaining redundant edges, called double
edges, belong to degree 2 vertices in $\dual{G}$ (e.g.  $\varphi_1^*(13)$,
$\varphi_1^*(14)$, $\varphi_1^*(15)$) in Fig.~\ref{fig:grid}b) and are removed
using a removal kernel of empty double edge (RKEDE) which contains all
darts incident to a degree 2 dual vertex. 
Further details about the construction scheme of a combinatorial
pyramid may be found in~\cite{brun-06-1}

As mentioned in Section~\ref{subsec:sampling}, if the initial
combinatorial map encodes a planar sampling grid, the geometrical
embedding of each initial dart corresponds to an oriented linel.
Moreover, each dart of a reduced map that is not a self loop encodes a
connected boundary between two regions. The embedding of the boundary
associated to such a dart may be retrieved from the embedding of the
darts of the initial map $G_0$. Let us consider the reduced
combinatorial map $G_i=(\dartset_i,\sigma_i,\alpha_i)$ defined at level $i$ and
one dart $d\in \dartset_i$ which is not a self loop. The sequence
$d_1\dots,d_n$ of initial darts encoding the embedding of the dart $d$
is obtained from the receptive field of $d$~\cite{brun-06-1} within
$G_0$ using the following relation:

\begin{equation}
\label{eq:def_seq}
d_1=d\;, d_{j+1}=\underset{m_j\mbox{ times}}{\underbrace{\varphi_0\circ\dots\circ\varphi_0}}(\alpha_0(d_j))
\end{equation}
where $\dual{G_0}=(\dartset_0,\varphi_0,\alpha_0)$ is the dual of the initial
combinatorial map and $m_j$ is the smallest integer $q$ such that
$\varphi_0^q(\alpha_0(d_j))$ survives at level $i$ or belongs to some former
RKEDE. The dart $d_n$ is the first dart defined by \Equ{eq:def_seq}
which survives up to level $i$. This dart also satisfies
$\alpha_0(d_n)=\alpha_i(d)$ by construction of the receptive fields. Note that
the tests performed on $\varphi_0^{q}(\alpha_0(d_j))$, $q\in \{1,\dots,m_j\}$ to
determine if it is equal to $d_{j+1}$ or $d_n$ are performed in
constant time using the implicit encoding of combinatorial
pyramids~\cite{brun-06-1}.

\subsection{Embedding of region boundaries}
\label{subsec:boundary}
Let us consider the dart $16$ in Fig.~\ref{fig:grid}c.  This dart
encodes the border between the background and the first row of the $3\times
3$ grid encoded by the $\sigma_3$ cycle $\sigma_3^*(16)=(16,7,8)$ of $G_3$. The
sequence of initial darts encoding the boundary of the dart $16$ is
retrieved using \Equ{eq:def_seq} and is equal to: $16.15.14.13.24$
(Fig.~\ref{fig:grid}b). We have for example $15=\varphi_0(\alpha_0(16))=\varphi_0(-16)$
(Fig.~\ref{fig:grid}c). Since each initial dart is associated to an
oriented linel, one may associate a sequence of Freeman's code to each
sequence of initial darts (Fig.~\ref{fig:grid}b) and thus to each dart
of a reduced combinatorial map $G_i$. The sequence of Freeman's codes
associated to a dart $d$ is denoted $s_d$ and is called the {\em
  segment} associated to $d$. for example, the segment associated to
the dart $16$ is equal  to $s_{16}=1.2.2.2.3$.


\section{Discrete geometry over a partition}
\label{sec:geometry}

\begin{figure}[t]
  \centering
  (a)~{\epsfig{file=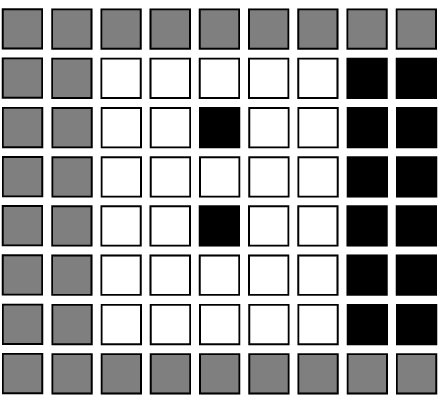,width=2.5cm}} 
  \hfill
  (b)~{\epsfig{file=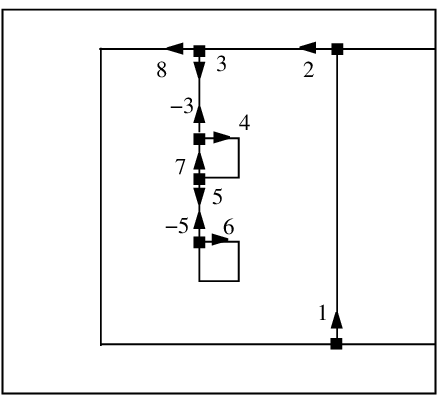,width=2.5cm}}
  \hfill
  (c)~{\epsfig{file=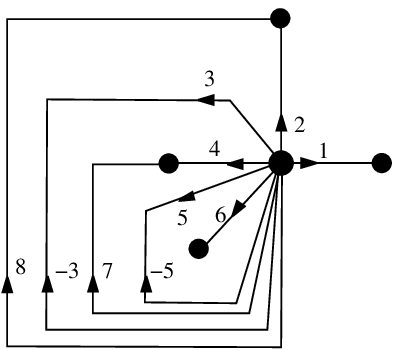,width=2.5cm}} 
  \vspace{-2mm}

  \caption{The central white region $\sigma^*(1)$ (a) contains several
    subregions. Its boundary is thus split into several connected
    components connected by bridges in $\dual{G}$ (b). These edges
    correspond to self loops in $G$ (c).}
  \label{fig:inclusion}
\end{figure}

As mentioned in Section~\ref{sec:combi_pyr}, each edge
$(d,\alpha_i(d))$ of a partition $G$ that is not a self loop encodes
a connected boundary between two regions. The edge is called {\em
separating}. On the other hand, a self loop corresponds to a bridge in
the dual combinatorial map and is characterized by $\alpha_i(d)\in
\sigma^*_i(d)$ (e.g. edge $(3,-3)$ or $(5,-5)$ in
Fig.~\ref{fig:inclusion}bc). Such edges, called {\em fictive}, 
either connect the outer boundary to some inner boundary (e.g. edge
$(3,-3)$ in Fig.~\ref{fig:inclusion}) or connect two inner boundaries
(edge $(5,-5)$ in Fig.~\ref{fig:inclusion}) \cite{brun-06-1}.

Each separating edge is embedded as a 4-connected digital path, included in the
interpixel digital plane (Section~\ref{sec:combi_pyr}.3 and
\cite{brun-06-1}).
When estimating the geometry of the boundary of the region,
fictive edges do not play any role. More precisely the concatenation
of only the separating edges defines also a set of 4-connected digital
loops. Each of these loops is either the outer boundary of the region
or one of its inner boundaries \cite{brun-06-1}. Given an initial
dart $d$ belonging to a separating edge, Algorithm~\ref{algo1} extracts a
boundary between region $\sigma^*(d)$ and its complement (setting
$L_\text{in}=\sigma^*(d)$) or between regions $\sigma^*(d)$ and
$\sigma^*(d')$ and their complement (setting
$L_\text{in}=\sigma^*(d)\cup \sigma^*(d')$). Its principle is to
follow the boundary with $\sigma$ except it skips fictive edges and
edges in-between $\sigma^*(d)$ and $\sigma^*(d')$. This method for
tracking a boundary is easily understood on Fig.~\ref{fig:inclusion}b,
where for instance the algorithm tracks from dart $1$, then
$2=\sigma(1)$, 3 is skipped since $-3 \in \sigma^*(1)$, then
$8=\sigma(-3)$ and terminates on $1=\sigma(8)$ again. Extracting all
the boundaries of a region is done in a similar way. All these
algorithms can be implemented with a complexity linear with the number
of boundary linels.

\begin{algorithm}[t]
\begin{algorithmic}
\item[1] Function Map::boundary( dart $d$, darts $L_{in}$ ) : Freeman chain \\
  \ENSURE{Return a sequence of Freeman's codes that is a 4-connected loop.} \\
  \REQUIRE{$d \not\in L_{in}$} \\
\item[2] 
  list $\DC \Assign \emptyset$,~~~dart $b$ \Assign $d$ \\
  \REPEAT
\item $\DC$.append($s_{b}$) \\
  $b$ \Assign $\sigma(b)$
  \WHILE[Skip fictive or interior edges]{$\alpha(b) \in L_{in}$}
\item $b$ \Assign $\varphi(b)$
  \ENDWHILE \\
\UNTIL $b = d$ \\
  {\bf return} $\DC$
\end{algorithmic}
\caption{Algorithm to visit all the linels of the digital boundary
  encircling region(s) specified by their darts $L_{in}$ and containing
  the dart $d$.
\label{algo1}}
\end{algorithm}

\subsection{Geometry with digital straight segments}  

We may now examine how geometric quantities can be estimated on a
closed 4-connected digital contour $C$, which is some boundary of a
region or two adjacent regions (computed as in the previous
paragraph). The literature is abundant on this topic and we restrict
ourselves to pure discrete geometry tools based on {\em digital
straight segment} (DSS) recognition. Several equivalent definitions of
DSS exist together with several classes of algorithms to recognize
them on digital curves (see for instance \cite{Klette04} for a recent
survey). We chose here to present briefly the arithmetic point of view
of digital lines, which leads to rather simple and efficient
algorithms \cite{Debled95,Feschet99}.


The set of points $(x,y)$ of the digital plane verifying $\mu \leq
ax-by < \mu + |a|+|b|$, with $a$, $b$ and $\mu$ integer numbers, is
called the {\em standard line} with slope $a/b$ and shift $\mu$.
A standard line is always 4-connected.  A sequence of consecutive
points $\PTS{i}{j}$ indexed from $i$ to $j$ of the digital curve $\DC$
is a {\em digital straight segment (DSS)} iff there exists a standard
line $(a,b,\mu)$ containing them. The one with smallest $a+b$
determines its {\em characteristics}, in particular its slope
$a/b$. Any DSS $Z$ thus defines an angle $\DD{Z}$ between its carrying
standard line and the x-axis (in $\lbrack 0; 2\pi \lbrack$ since a DSS
is oriented), called the {\em direction of $Z$}.

\begin{figure}[t]
  \begin{center}
    \epsfig{width=0.33\textwidth,file=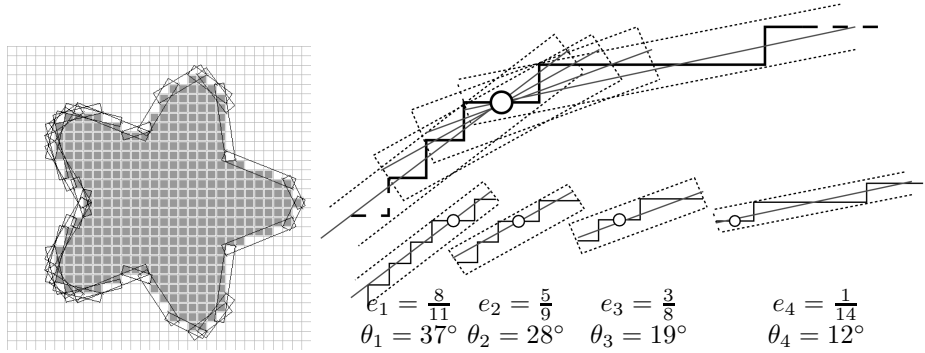}
    \input{MS-tangentes.pstex_t}
  \end{center}
  \vspace{-0.3cm}
  \caption{Left: every maximal segment along this contour is drawn as
    its rectangular bounding box. Right: $\lambda$-Maximal Segment
    tangent estimation at a given point.
    \label{fig:tcover}}
\end{figure}

The predicate {\em ``$\PTS{i}{j}$ is a DSS''} is denoted by
$\SPRED{i}{j}$. Incremental algorithms exist to recognize a digital
straight segment on a curve and to extract its characteristics
\cite{Debled95}. Therefore deciding $\SPRED{i}{j+1}$ or
$\SPRED{i-1}{j}$ from $\SPRED{i}{j}$ are $O(1)$ operations. Any DSS
$\PTS{i}{j}$ is called a {\em maximal segment} iff $\lnot
\SPRED{i}{j+1}$ and $\lnot \SPRED{i-1}{j}$. Maximal segments are thus
the inextensible DSS of the curve (Fig.~\ref{fig:tcover}, left). Note
that the set of all maximal segments of a curve can be computed in
time linear with the number of curve points \cite{Feschet99}.

\subsection{Tangent, normal and length estimation}
\label{subsec:tangent}
\begin{sloppypar}
Several tangent estimators based on DSS recognition have been proposed.
We propose to use the $\lambda$-Maximal Segment Tangent estimator
($\lambda$-MST) to approach the tangent direction at any point of the
digital curve \cite{Lachaud05a}. It was indeed shown to give good
approximations even at coarse scale, to be rather independent from
rotations and to be asymptotically convergent.

Fig.\ref{fig:tcover}, right, gives the essential idea of this tangent
estimator. Given a point, the direction $\theta_i$ of every maximal
segment containing it is evaluated. The relative position $e_i$ of the
point within the maximal segment is also computed. The $\lambda$-MST
tangent direction $\hat{\theta}$ is some weighted combination of the
preceding parameters: $\hat{\theta}=(\sum_i
\lambda(e_i)\theta_i)/(\sum_i \lambda(e_i))$. In our experiments, the
mapping $\LF$ was defined as the triangle function taking base value 0
at 0 and 1, and peak value 1 at $\frac{1}{2}$.  For further details,
see \cite{Lachaud05a}.
\end{sloppypar}


The experimental average number of maximal segments per linel is
between 3 and 4. Therefore computing the $\LF$-MST direction is not
costly and is a $O(1)$ operation on average. This technique of tangent
estimation is easily extended to any real curvilinear abscissa along
the digital contour. The tangent is thus defined at any linel, taking
half integer abscissas.

The estimation of the {\em normal vector} at $\PT{k}$ is then simply
the vector $\NORMAL{k}=(-\sin(\LTO{k}),\cos(\LTO{k}))$.  The {\em
elementary length} $\LE{k,k+1}$ of a linel $\PTS{k}{k+1}$ is defined
as $|\cos(\LTO{k+0.5})|$ for horizontal linels and
$|\sin(\LTO{k+0.5})|$ for vertical linels. It corresponds to an
estimation of the length of a unit displacement along the digital
curve.  The length of $\DC$ is estimated by simple summation of the
elementary length of its linels. This method of length evaluation was
reported to give very good experimental results \cite{Coeurjolly04}.
If $\DC^b$ is boundary($b,\sigma^*(b)$) as returned by
Algorithm~\ref{algo1}, then its {\em length} is
$\LENGTH{b}{G}=\sum_{k=1}^{|\DC^b|} \LE{k,k+1;\DC^b}$. The total {\em
perimeter} $\PERIM{R(\sigma^*(b))}{G}$ of the region $\sigma^*(b)$ is
the sum of the length of each of its boundaries.



\section{Energy of a partition and pyramidal segmentation}
\label{sec:energy}

The geometrical features (normal, perimeter, polygonalization) defined
in Section~\ref{sec:geometry} may be computed on each region of a
partition in order to provide different measures of its geometrical
characteristics. Such measures may then be incorporated into a
hierarchical segmentation algorithm based on an energy minimization
scheme (Section~\ref{sec:intro}). Such energy balances two terms: the
goodness of fit term and a regularization term which penalizes
unlikely or complex models. The energy of a partition encoded by the
map $G$ is simply called the {\em energy of the combinatorial map $G$}
and is formally defined as follows: Let $G=(\dartset,\sigma,\alpha)$ be a
combinatorial map with a geometrical embedding in the digital grid and
an input image $I$ over this grid. Let $\dartset_\sigma$ be the set of
$\sigma$-cycles of $\dartset$. The energy of the combinatorial map $G$ is

\vspace{-1mm}
\noindent
\begin{minipage}[t]{0.41\linewidth}
  \begin{equation}
    \E(G) = \sum_{\sigma^*(d) \in \dartset_\sigma} \E(\sigma^*(d))
    \label{eq:eng-partition}
  \end{equation}
\end{minipage}
\hfill
\begin{minipage}[t]{0.6\linewidth}
  \begin{equation}
    \E(\sigma^*(d))=\Eimg(\sigma^*(d))+ \nu \Ereg(\sigma^*(d)) \label{eq:eng-region}
  \end{equation}
\end{minipage}

\Equ{eq:eng-partition} indicates that the global energy is
decomposable over each region. This property helps in defining fast
algorithms for region decimation. \Equ{eq:eng-region} balances the two
energies, one dependent on the image (the image energy \Eimg), the
other dependent only on the model (the regularization energy \Ereg).

The parameter $\nu$ is often interpreted as a scale parameter, since it
privileges the goodness of fit for low values (and over-segmentation)
and {\em a priori} most likely regions for high values (and
under-segmentation).

The image energy used within our experiments is defined as follows:

\begin{equation}
  \Eimg(\sigma^*(d))= - \delta \sum_{\PT{k} \in \DC^d} \|D~I(\PT{k})\| \hat{l}(k,k+1)  + {\sum_{(x,y) \in R(\sigma^*(d))} \|I(x,y) -\mu_{\sigma^*(d)}\|^2}\label{eq:Eimg}
\end{equation}

where $\hat{l}(k,k+1)$ denotes the length estimate of a lignel at point
$k$, $I(x,y)$ denotes the color of the pixel $(x,y)$ and
$\|D~I(\PT{k})\|$ the norm of the differential of $I$ at point $k$. This
last measure is equal to the norm of the gradient for grey level
images.  The term $\mu_{\sigma^*(d)}$ represents the mean color of the region
encoded by $\sigma^*(d)$. The second sum of the above expression denotes
thus the squared error of the region. Finally, the term $\delta$ represents
the respective weight of the gradient and squared error energies.

\begin{figure}[t]
  \begin{center}
    \begin{tabular}{cccc}
      \epsfig{height=3cm,file=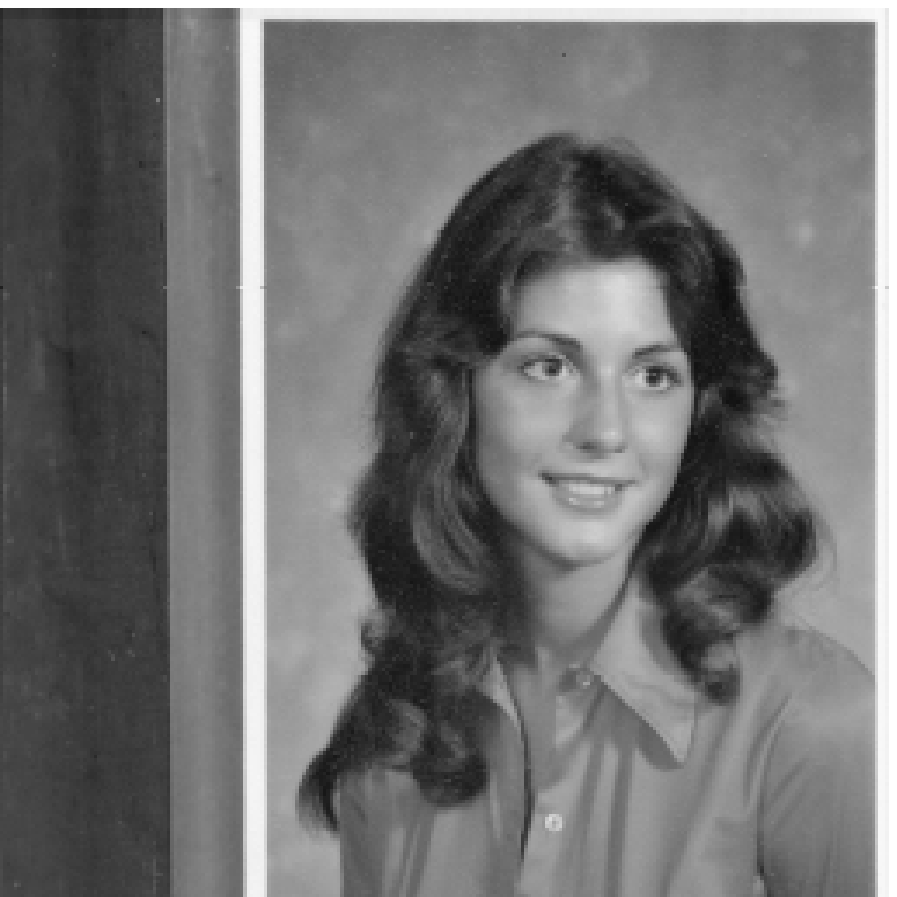}&
      \epsfig{height=3cm,file=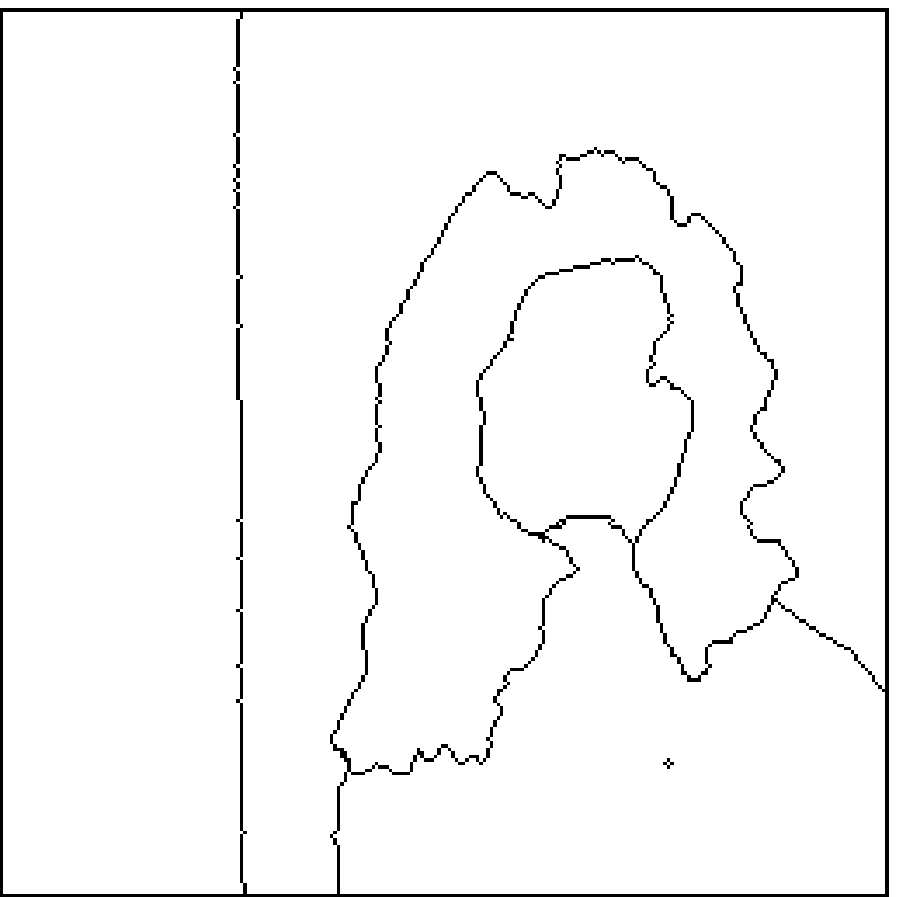}&
    \epsfig{height=3cm,file=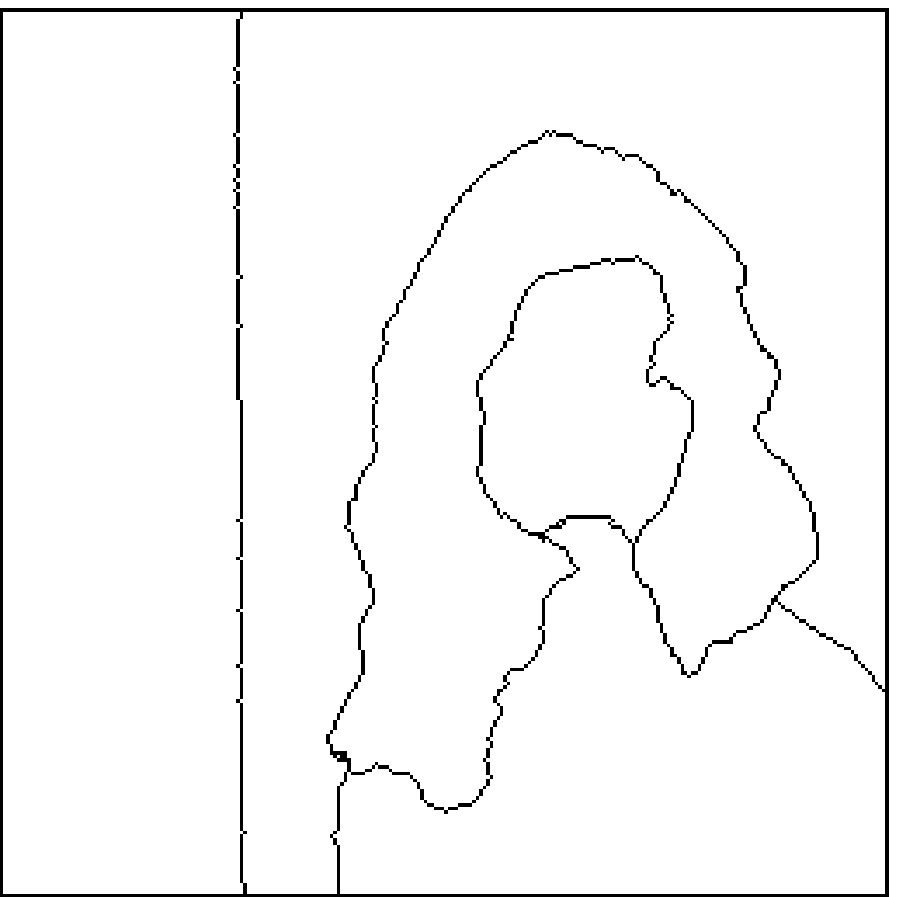}&
    \epsfig{height=3cm,file=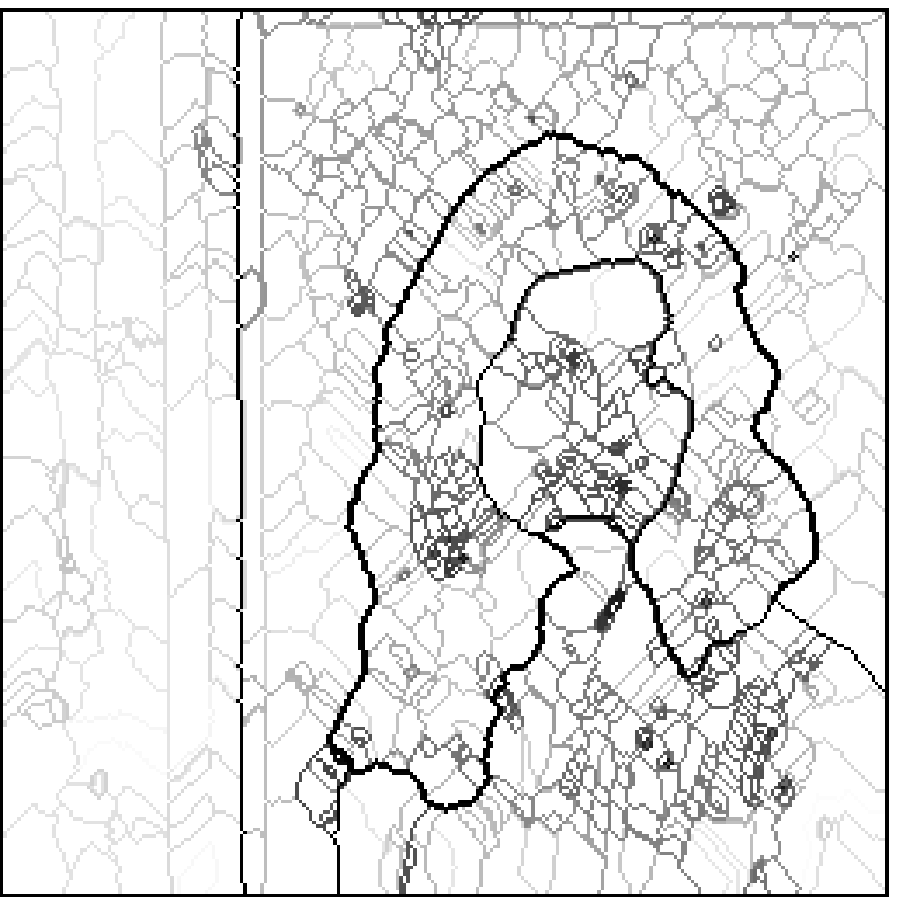}\\
    (a)&(b)&(c)&(d)\\
  \end{tabular}
  \end{center}
  \caption{Influence of length penalization: (a) image Girl, (b) one level
of the pyramid with $\hat{l}(k,k+1)=1$, (c) same level of a pyramid built using   discrete length estimators. All the boundaries of the pyramid which contains (c) are superimposed on (d). The darkest boundaries are those who survive at the highest levels.
%
    \label{fig:girl}}
\end{figure}

 
The regularization energy is defined from the estimate of the
perimeter of the region as $\Ereg(\sigma^*(d))=
\PERIM{R(\sigma^*(d))}{G}$(Section~\ref{subsec:tangent}). Given two
possible merge operations inducing the same variation of the image
energy, this choice for the regularization term favors the one which
induces the simplest partition with the lowest overall length of
contours. The advantage of using discrete length estimators compared
to a basic count of the lignels is to make the segmentation process
more independent on the alignment of components wrt some axes.

We tested the influence of length penalization on the classical Girl
test image (Fig.~\ref{fig:girl}). Two pyramids have been built on an
initial partition encoded by a combinatorial map $G_0$. This partition
is defined by a watershed algorithm applied on the gradient of the
Girl test image. The parameter $\nu$ is fixed to $1.3$ during the
construction of both pyramids. Fig.~\ref{fig:girl}(b) represent one
level of the first pyramid built using a fixed length estimate equal
to 1 for all lignels.  Fig.~\ref{fig:girl}(c), represents the same
level within the second pyramid built using the discrete length
estimator defined in Section~\ref{subsec:tangent}. As shown by
Fig.~\ref{fig:girl}(c) the more accurate measure of the length given
by the discrete length estimator provides smoothest boundaries.

\subsection*{Pyramidal segmentation algorithm}

Our energy minimisation method starts with an initial partition coded
by a map, and merges at each step the two adjacent regions, the
merging of which induces the greatest decrease (or the smallest
increase) of the combinatorial map energy. This process may be
interpreted as a gradient descent which continues when a local minima
is reached in order to seek for other minima.  Note that our framework
is not devoted to a specific strategy for energy minimization.  Many
alternative optimization heuristics could be used (e.g. the
scale-climbing of Guigues {\Etal} \cite{guigues-06}). The proposed
approach is however sufficient to compare the respective advantages of
different energies. Let us additionally note that using our strategy
or the scale climbing of Guigues et al., only two regions are merged
between two consecutive levels of the pyramid. This merge strategy
does not induce a high memory cost due to the implicit encoding of the
combinatorial pyramid~\cite{brun-06-1}. An explicit construction of
all the reduced graphs using graph or dual graph pyramids would
require a huge amount of memory with a lot of redundancy between
graphs.




\section{Conclusion}

We have presented a new framework for segmenting images with a
pyramidal bottom-up approach using an energy-minimizing scheme.  Our
framework combines combinatorial pyramids, which can represent in the
same structure all the levels of a hierarchy, and discrete geometric
estimators, which provide precise geometric measurements and allow the
definition of new regularization and image energy terms. A greedy
algorithm for computing the hierarchy was also provided and some
examples of segmentation were exhibited and discussed.

Our first experiments show that, the length estimation can have a
great influence on the regularization of the segmentation. Discrete
geometric estimators provides some smoothest boundaries. However, they
are useless if the over-segmentation gives irregular regions.  In
futur works, we want to tackle this problem by using a smoothest
over-segmentation.



\bibliographystyle{splncs}
\bibliography{biblio-jol,comp-graph}

\end{document}